# A Natural Language Processing Approach to Support Biomedical Data Harmonization: Leveraging Large Language Models

Zexu Li[a], MS, Suraj P. Prabhu[b], MS, Zachary T. Popp[a], MPH, Shubhi S. Jain[c], MPH, Vijetha Balakundi[d], MS, Ting Fang Alvin Ang[a,c,e], MD, MPH, Rhoda Au[a,c,e,f,g,h], PhD, Jinying Chen[d,i*], PhD

[a] Department of Anatomy and Neurobiology, Boston University Chobanian & Avedisian School of Medicine, Boston, MA, USA

[b] Department of Bioinformatics, Boston University Faculty of Computing & Data Sciences, Boston, MA, USA

[c] Slone Epidemiology Center, Boston University Chobanian & Avedisian School of Medicine, Boston, MA, USA

[d] Department of Medicine/Section of Preventive Medicine and Epidemiology, Boston University Chobanian & Avedisian School of Medicine, Boston, MA, USA

[e] Framingham Heart Study, Boston University Chobanian & Avedisian School of Medicine, Boston, MA, USA

[f] Department of Neurology, Boston University Chobanian & Avedisian School of Medicine, Boston, MA, USA

[g] Department of Epidemiology, Boston University School of Public Health, Boston, MA, USA

[h] Department of Medicine/Section of Genetics, Boston University Chobanian & Avedisian School of Medicine, Boston, MA, USA

[i.] Data Science Core, Boston University Chobanian & Avedisian School of Medicine, Boston, MA, USA

# Abstract

**Objective:**

Biomedical research requires large, diverse samples to produce unbiased results. Retrospective data harmonization is often used to integrate data from existing datasets or biobanks to create these samples, but the process is labor-intensive. Automated methods for matching variables across datasets can accelerate this process. Research in this area has been limited, primarily focusing on lexical matching and ontology-based semantic matching. We aimed to develop new methods, leveraging large language models (LLM) and ensemble learning, to automate variable matching.

**Methods:**

We utilized data from two GERAS cohort (European [EU] and Japan [JP]) studies obtained through the Alzheimer's Disease (AD) Data Initiative's AD workbench to develop variable matching methods. We first manually created a dataset by matching 352 EU variables with 1322 candidate JP variables, where matched variable pairs were positive and unmatched pairs were negative instances. Using this dataset, we developed and evaluated two types of natural language processing (NLP) methods, which matched variables based on variable labels and definitions from data dictionaries: (1) LLM-based, including E5, MPNet, and MiniLM; and (2) fuzzy matching. We then developed an ensemble-learning method, using the Random Forest (RF) model, to integrate individual NLP methods. RF was trained and evaluated on 50 trials. Each trial had a random split (4:1) of training and test sets, with the model's hyperparameters optimized through cross-validation on the training set. For each EU variable, 1322 candidate JP variables were ranked based on NLP-derived similarity scores or RF's probability scores, denoting their likelihood to match the EU variable. Ranking performance was measured by top-n hit ratio (HR-n) and mean reciprocal rank (MRR).

**Results:**

E5 performed best among individual methods, achieving 0.90 HR-30 and 0.70 MRR. RF performed better than E5 on all metrics over 50 trials ($P<0.001$) and achieved an average HR-30 of 0.98 and MRR of 0.73.

LLM-derived features contributed most to RF's performance. One major cause of errors in automatic variable matching was ambiguous variable definitions within data dictionaries.

**Conclusion:**

NLP techniques (especially LLMs), combined with ensemble learning, hold great potential in automating variable matching and accelerating biomedical data harmonization.

## 1. Introduction

Epidemiology and machine learning studies in the biomedical domain often require large, diverse samples to produce unbiased results and improve the generalizability of findings [1, 2]. However, such comprehensive data are rarely found in a single study. Instead, many datasets are generated by individual studies and shared via public platforms or data repositories [3]. Data sharing has become widely adopted in research communities and is now often mandated by funding agencies [4, 5]. To effectively utilize the shared datasets, data harmonization is typically employed or required [6-8].

Data harmonization refers to the process of combining data from multiple resources to achieve maximum compatibility [6]. Strategies for data harmonization can be broadly categorized into two types: the stringent approach and the flexible approach [9]. The stringent approach uses the same measurements and data collection protocols across studies. Datasets from these studies share the same variables and can be harmonized through simple data merging [10]. In practice, implementation of this strategy is often challenging and typically confined to specific projects, due to the absence of widely accepted common data elements (i.e., standardized data definitions) and low adoption rates across individual studies[11]. In addition, the availability of numerous historical datasets necessitates effective harmonization methods that are both flexible and robust to maximize their usability[8]. The flexible approach does not require studies to use identical variables; instead, it uses analytical methods to transform matched variables into a common data model [12]. For example, the flexible prospective harmonization method requires researchers to agree on compatible data collection tools and protocols before the studies begin [13]; while the flexible retrospective harmonization method can be applied to existing data of similar studies without prior collaboration or agreement among these studies [14, 15].

All flexible retrospective harmonization methods share a common, early step: identifying variables that can be merged or mapped across studies (which we call variable matching) [8, 16]. Current data harmonization practices in the biomedical domain tackle this problem in manual and labor-intensive ways, where cohort experts identify relevant variables, assess their compatibility (based on the variable definition and data format), and estimate the level of difficulty of harmonization [8, 14, 16]. Furthermore, even studies using similar data collection protocols may use different variable naming conventions [17-19]. Approaches that automatically match variables across studies can reduce manual efforts and accelerate the data harmonization process. Research in this area has been limited, with methods primarily focusing on lexical matching and ontology-based semantic expansion and matching (e.g., using concepts and relations defined in ontologies to expand variables to be matched) [20-23].

This study aimed to develop and evaluate new variable matching methods by leveraging large language models (LLMs) to compare lexical and semantic similarities of variable labels and definitions across studies. We further trained a Random Forest classifier to ensemble features generated by the individual natural language processing (NLP) methods, including fuzzy matching and LLMs, to match variables. The evaluation of these methods on matching variables from two cohort studies, the GERAS-Europe (GERAS-EU) [17] and GERAS-Japan (GERAS-JP) [18] studies, showed that the LLMs achieved decent results, and the ensemble method performed consistently better than the individual NLP methods in matching variables.

## 2. Methods

### 2.1 Approach overview

We treated the variable matching task as a ranking problem (Figure 1). For each source variable (i.e., a variable from GERAS-EU), we ranked candidate target variables (e.g., variables from GERAS-JP) based on their similarities to the source variable. The similarity between two variables were estimated using information extracted from variable labels and other related sources (e.g., definitions or derivation rules of the variables, descriptions of data sheets containing the variables) by individual NLP methods and ensemble learning. Figure 1 provides an overview of our approach.

### 2.2 Study setting

We mapped variables between the GERAS-EU [17] and GERAS-JP [18] studies, which were accessed through the Alzheimer's Disease (AD) Data Initiative's AD Workbench — a secure, cloud-based data sharing and analytics environment to facilitate open data access and collaboration for AD related research globally. The GERAS-EU study is an observational study that investigated the societal costs associated with AD in three European countries: France, Germany, and the United Kingdom [17]. This study recruited 1,532 participants with probable AD between October 1, 2010, and September 31, 2011, and collected data during the baseline visit and the follow-up visits every 6 months over 18 to 36 months. A variety of data elements were collected, such as demographics, medical history, cognitive function (e.g., Mini-mental state examination [MMSE] [24], the Alzheimer's Disease Assessment Scale–Cognitive Subscale [25]), daily activities (Activities of Daily Living Inventory [26]), resource utilizations (e.g., Resource Utilization in Dementia survey [27]), and quality of life (e.g., EuroQol-5 Dimension surveys [28]). The second study, GERAS-JP, utilized a similar study protocol to investigate the societal costs associated with AD in Japan

[18, 29]. It enrolls 553 participants with probable AD between November 2016 and December 2017 in Japan. Although GERAS-JP and GERAS-EU collected similar data, the way they recorded and formatted their data differed. As shown in Table 1 (see Appendix 1, Table A1-1, for more examples), the variable recording the participants' body mass index was named "BMIB" (labeled as "Body Mass Index (BMI) at Baseline") in GERAS-JP and "VSBLVTR_BMI" (labeled as "Vital Sign Result Numeric BMI baseline") in GERAS-EU. Both variables represent body mass index, but their names and variable labels are different.

## 2.3 Data preprocessing and information used for variable matching

We conducted long-to-wide transformation on the data before matching variables. For example, MMSE scores in GERAS-EU were stored in long format, where the variable "MMSEQSNUM" recorded the MMSE question number, and "MMSERN" recorded the corresponding scores. In contrast, MMSE scores in GERAS-JP were stored in wide format, with variables such as "MMSE_Q1" and "MMSE_Q2" recording scores for each MMSE question separately. We transformed "MMSEQSNUM" and "MMSERN" jointly into a wide format and assigned variable labels to the transformed variables accordingly. The same process was applied to all questionnaire-related variables originally represented in long format. Below, we describe the information we extracted from the data dictionaries for variable matching.

### 2.3.1 Variable labels and data sheet description

Table 1 provides examples of matched variables from the GERAS-EU and GERAS-JP studies (see Table A1-1 in Appendix 1 for more examples). Variable labels provide descriptions of the variable names. Both studies organized their variables in separate data sheets, with each data sheet representing a specific type of variables (e.g., demographic variables) or variables derived from a specific survey. Each data sheet was accompanied by a short description in the data dictionaries. For example, the GERAS-JP study kept variables associated with the MMSE in a data sheet with the description "Mini-Mental State Examination

(MMSE) per visit". In this data sheet, the variable "MMSE_Q8" recorded the response to the eighth question in the MMSE [24].

Table 1. Examples for variable names, labels, and data sheet descriptions [a]

| EU Variable Name | EU Variable Label | EU Data Sheet Description | JP Variable Name | JP Variable Label | JP Data Sheet Description |
|---|---|---|---|---|---|
| **Demographics** | | | | | |
| SEXLNM | Sex LNM | Demographics Relationship | SEX | Sex | Subject Level Analysis Dataset |
| **Anthropometric** | | | | | |
| VSBLVTR_BMI | Vital Sign Result Numeric BMI baseline | Vitals | BMIB | Body Mass Index (BMI) at Baseline | Subject Level Analysis Dataset |
| **Diagnosis/Treatment** | | | | | |
| DISCDLNM_Hypertension | Disease Code SNM: Hypertension | Medical History - Comorbidities | CCC09 | Hypertension Diagnosis | Caregiver Comorbidities per visit |
| **Questionnaire** | | | | | |
| MMSERN_MMSES8 | MMSE Item Result Numeric MMSES8 Correct response to orientation to place what is the city/town | Mini-Mental State Exam | MMSE_Q8 | What is the city/town? | Mini-Mental State Examination (MMSE) per visit |
| **Time to event** | | | | | |
| TTERN_TTD | Time to Event Result Numeric (Time) TTD: Time to Death | Time to Event | TTDEATH | Time to Death (Months) | Subject Level Analysis Dataset |
| **Cost** | | | | | |

| COSPR1RN | Cost Primary Analysis 1 Item Result Numeric | Cost Caregiver Indirect Nonmedical | COST_INC_OP_C24_SUM | Caregiver Indirect Non-Medical Cost from Baseline up to the Visit: Opportunity Cost Approach, Supervision Time Not Included, Average Hours p.d. capped on 24 | Total Cost up to the visit per visit |
|---|---|---|---|---|---|

ᵃ Each row in the table are two matched variables from GERAS-EU and GERAS-JP studies

### 2.3.2 Key word extraction from derivation rules

The data dictionaries of both GERAS-EU and GERAS-JP studies also contain a field called variable definition or derivation rule (see Table A1-2 in Appendix 1 for examples). This field provides information about how the variables were defined or derived, such as the values of categorical variables and the derivation rules of variables. Derivation rules provide additional information beyond variable labels and data sheet descriptions, which can be valuable for variable matching. However, not all variables have derivation rules. In addition, there is significant variation in length and content of derivation rules. Therefore, we utilized information from derivation rules exclusively for ensemble learning.

We incorporated information from derivation rules into variable matching as follows. First, if a derivation rule contained more than 20 words, KeyBERT [30] was employed to extract key words of up to 15 words to represent the entire text. KeyBERT utilizes BERT (Bidirectional Encoder Representation from Transformer) -based embeddings to identify sub-phrases in a text that are most similar to the original text. If the derivation rule contained 20 or fewer words, the entire text was retained. Second, for each variable, we concatenated the variable label with the derivation rule (or key words extracted from the derivation rule) to create input for the individual NLP methods. This treatment ensured that variables without derivation rules had non-empty input for NLP. The outputs from the individual NLP methods were then used as machine learning features for ensemble learning.

## 2.4 Natural language processing methods for variable matching

We evaluated two types of NLP methods: LLM-based and fuzzy matching.

### 2.4.1 Large language model-based methods

We evaluated three LLMs in variable matching: E5, MPNet and MiniLM.

The E5 (**E**mb**E**ddings from bidir**E**ctional **E**ncoder R**E**presentations) is a text embedding model that enhanced its training process through weakly supervised contrastive pre-training [31]. The key idea of E5's contrastive pre-training was optimizing text embeddings so that they would bring relevant unlabeled text pairs closer together and push irrelevant text pairs further apart within the vector space of embeddings [32]. In E5, relevant text pairs were sourced from diverse platforms including Reddit (posts and comments), Stack Exchange (questions and upvoted answers), English Wikipedia (entity names and passages), Scientific papers (titles and abstracts), and Common Crawl web (titles and passages) and selected via a consistency-based data filtering technique [31]. These relevant text pairs served as positive examples; text from different relevant pairs formed irrelevant text pairs (i.e., negative examples). During pre-training, E5 enhanced existing LLMs, e.g., the BERT models by leveraging the large amount of newly collected text pairs and contrastive learning. The model was then fine-tuned using three labeled datasets: NLI (Natural Language Inference), MS-MARCO passage ranking dataset [33], and NQ (Natural Questions) [34] datasets. E5 outperformed existing embedding models on both BEIR [35] and MTEB [36] benchmark datasets that were used to evaluate a variety of text embedding tasks. This study used the E5_large model, which is built on the large BERT model Bert-large-uncased-whole-word-masking model. The E5_large_v2 model was chosen due to its superior performance on benchmark datasets, particularly in the STS (Semantic textual similarity) tasks [31].

In addition, we evaluated two LLMs developed using the Sentence Transformers (also called SBERT) framework [37]. Built on Siamese and triplet networks and contrastive learning techniques, SBERT aims to generate semantically meaningful embeddings from sentences or short paragraphs while achieving higher computational efficiency than BERT [37]. SBERT incorporates pre-trained LLMs into their low-level building blocks. We evaluated two LLMs, MPNet and MiniLM, incorporated into SBERT [38]. The MPNet (masked and permuted language modeling) model unifies mask language modeling from BERT and permuted language modeling from XLNet [39]. In addition, MPNet utilizes auxiliary position information (i.e., the tokens' positions in the original, non-permutated input sentence) to improve the consistency of model's input representations between pre-training and fine-tuning [39]. The All_MPNet_base model is fine-tuned on the MPNet-base model using contrastive learning with over 1 billion text pairs sourced from diverse datasets (e.g., Stack-Exchange, MS-MARCO, NQ) [38]. Among all SBERT models, the All_MPNet_base model demonstrates superior performance on Sentence Embedding Task (14 datasets) and Semantic Search Task (6 datasets) [38].

The MiniLM model employs a specific knowledge distillation technique, deep self-attention distillation, to compress large Transformer-based models [40, 41]. Knowledge distillation compresses a large teacher model into a smaller student model with fewer parameters by minimizing the difference between features of the two models (e.g., self-attention distributions in Transformer models). It has been shown that the student model obtained through using this technique can maintain similar test accuracy as the teacher model across tasks such as image and speech recognition [42]. MiniLM employs distillation on the self-attention module of the final transformer layer of the BERT base model, resulting in a 12-layer student model [41, 43]. This reduction in parameters enhances fine-tuning efficiency. In this study, we used All_MiniLM_L12 model [37, 38]. Initialized from the pre-trained model MiniLM (microsoft/MiniLM-L12-H384-uncased), All_MiniLM_L12 was fine-tuned with a contrastive objective using a dataset of 1 billion text pairs, including NQ and SQuAD2.0 [44].This model is selected due to its comparable performance in sentence embedding and semantic search tasks when compared with the All_mpnet_base_v2 model [38].

### 2.4.2 Fuzzy matching

Fuzzy matching methods utilize dynamic programming and edit distance [45](e.g., Levenshtein distance) to assess similarities between sentences. Edit distance measures the number of operations needed to transform one text string into another [45]. We computed the edit distance between the label of a GERAS-EU variable and the label of each GERAS-JP variable. Variables with a smaller edit distance were considered similar. We implemented the fuzzy matching method by using the Python package RapidFuzz [46], which incorporates a variety of scoring functions for edit distance that were originally developed in the Fuzzywuzzy Package [47]. Prior to applying fuzzy matching, we preprocessed the variable labels by removing punctuations and stop words, converting all letters to lowercase, and stemming the words. We compared performance of various fuzzy matching scoring functions implemented in the RapidFuzz package in a preliminary experiment and selected the top-performing function, the token-set-ratio function, for this study. The token-set-ratio function is an extension of the token-sort-ratio function. The token-sort-ratio function tokenizes the preprocessed strings, sorts the tokens alphabetically, and computes the Levenshtein distance between the sorted strings. The token-set-ratio function eliminates duplicate tokens within each string before comparison.

## 2.5 Ensemble learning for variable matching

To further improve variable matching performance, we employed the Random Forest classification algorithm to integrate the outputs from both LLM-based methods and the fuzzy matching method.

### 2.5.1 Random Forest classifier

The Random Forest classifier is a supervised machine learning model that utilizes an ensemble of decision tree classifiers to generate predictions. The classifier takes features associated with a pair of EU-JP variables as its input and outputs a class label (1: matched variables, 0: unmatched variables) and the associated probabilities. The training of the Random Forest classifier involves the creation of an ensemble of decision trees that classify the input from the training data. Each tree in the random forest builds on a subset of training instances randomly sampled from the complete training set without replacement, as well as a subset of features randomly selected from the entire feature set [48]. The introduction of randomness serves to enhance robustness and mitigate overfitting. The construction of a decision tree involves splitting nodes iteratively from top to bottom. Each node is split based on a specific machine learning feature, with the feature and corresponding splitting rule determined by criteria such as Gini impurity and information gain. For continuous features, the split rule may check whether the feature value is within a certain range. For categorical features, it may check whether the feature value is equal to a specific category. The goal is to achieve the greatest reduction in impurity or the largest increase in information gain with each split. The node-splitting process continues until certain termination criteria are met, such as reaching the maximum tree depth, the minimum number of samples required for a split, or the minimum decrease in impurity. When applying a trained Random Forest classifier to classify a data instance, each decision tree that makes up the classifier is applied to the data instance respectively. At each node of a decision tree, the corresponding split rule will direct the data instance to a certain branch (i.e., a specific child node) under that node. After traversing several nodes in the tree, the data instance will reach a leaf node and receive its classification label (which is the label of the majority training instances that reach the same leaf node during model training). A final classification label is determined based on the aggregated voting results from all decision trees. In addition, the classifier will output a probability denoting how likely the data instance is positive, i.e., a pair of matched variables. These probabilities were then used to rank the candidate JP variables for each EU variable. Figure 2 provides an overview of the training and test procedures of the Random Forest model.

### 2.5.2 Training and test datasets

The datasets utilized in this study were created in the following way.

First, we manually matched a subset of variables from GERAS-EU and GERAS-JP studies (i.e., the ground truth variable pairs). Several steps were taken to ensure the matching accuracy. At the first step, EU variables were assigned to five co-authors, who had training and work experiences in epidemiology, statistics, and bioinformatics, for preliminary matching with corresponding JP variables. The variables were matched based on both variable definitions (i.e., derivation rules) and variable values. Challenging cases were discussed within the research team, including a senior co-author with expertise in epidemiology and health informatics. We categorized the matching results into three groups: no match, single match, and multiple matches. For example, the variable "Sex" ("SEXLNM") in the GERAS-EU study was matched with multiple variables ("ADSL_SEXCD", "ADCOV_SEXCD") in the GERAS-JP study as it was included in multiple data sheets under different variable names in GERAS-JP. At the second step, the five co-authors validated and corrected matching results from the first step, with each assigned a different subset of variables. The validation results and corrections (along with a justification for any corrections) were documented. At the last step, one co-author reviewed all corrections to ensure their validity. Through this 3-step procedure, 447 pairs of EU-JP variables were manually matched, which included 352 unique EU variables (some of which have multiple true alignments).

Second, we created training and test datasets in three steps (Figure 2). First, we treated the 352 unique EU variables as source variables that need to be aligned with the target variable(s) in the GERAS-JP study and randomly divided these variables into training and test (4:1) sets. Each EU source variable had 1322 JP candidate variables to match, which constituted 1322 variable pairs. These variable pairs contained only one or few positive instances (matched EU-JP pairs) but many negative instances (unmatched EU-JP pairs). Second, we down sampled the negative instances to reduce the negative impact of data imbalance on model training. Specifically, for each EU source variable, we randomly selected 200 JP variables that did not match this EU variable to form 200 EU-JP variable pairs as negative training instances. In the test set, however, we included all the negative instances. This resulted in about 282 positive and 56,640 negative instances in the training set and 70 positive and 92,530 negative instances in the test set in each trial of the machine learning experiment (detailed in section 2.7 Experimental settings).

### 2.5.3 Class label and machine learning features

The class label is binary-valued, with 1 indicating manually matched EU-JP variable pairs and 0 indicating unmatched EU-JP variable pairs. Features utilized by the Random Forest model came from two sources: similarity scores generated by four individual NLP methods (E5, MPNet, MiniLM, and fuzzy matching) and other information extracted from data dictionaries. Each NLP method generated three similarity scores for an EU-JP variable pair by using data labels, data sheet descriptions, and key words extracted from derivation rules as the respective input. For example, the EU variable "DIAGDT" (representing "Disease diagnosis date") and the JP variable "ADDIADT" received three similarity scores from the E5 model calculated based on variable labels (0.98), data sheet descriptions (0.76), and key words from derivation rules (0.90) respectively. Other features include the length of variable labels, the length and the absence of derivation rules for both the EU and JP variables. For example, for the EU variable "DIAGDT", we had three features: 3 for the length of the variable label measured in words, 30 for the length of the derivation rule in words, and "False" for the absence of derivation rules. In total, 12 similarity scores (three generated by each NLP method) and 6 additional features (3 for each variable in the variable pair) were used as machine learning features.

### 2.5.4 Model development

In this study, we utilized the Random Forest classifier from the Scikit-learn package [49]. We developed the model using the training set and tuning the following hyperparameters through 5-fold cross validation and grid search: (1) the number of trees in the forest; (2) the maximum depth of each tree; (3) the criteria used to assess the quality of a tree node split; (4) the minimum number of samples required for each split; and (5) The number of features considered at each splitting. Other hyperparameters were set to their default values. We used the mean reciprocal rank (MRR; see details in section 2.6 Evaluation metrics) to select the best hyperparameters.

## 2.6 Evaluation metrics

For each EU variable, each matching method will generate a ranked list of all candidate JP variables. A high-performance method is expected to place the manually matched JP variable(s) at the top of this list. We evaluated the performance of the matching methods by two metrics: hit ratio (HR) and MRR. Both HR and MRR are commonly used to evaluate recommendation or ranking algorithms [50, 51]. As shown in equation 1, HR is defined as the total number of hits appearing in the top-n ranked items ($V_{hit}^n$) divided by the total number of search queries ($V_{all}$). A "hit" denotes that the user-selected item is among the top-n ranked items. In our case, a "hit" means, for an EU variable, the manually matched JP variable is within the top-n ranked list of JP variables. $V_{hit}^n$ denotes the total number of hits in the top-n lists for all EU variables in a dataset. $V_{all}$ denotes the total number of EU variables in the dataset. For each user query $v$, the reciprocal rank RR($v$) is the inverse of the rank of the first relevant item (equation 2). The MRR is the averaged reciprocal rank across all queries (equation 3). In our case, $R_{hit}$ in equation 2 denotes the highest (smallest) rank of the JP variable(s) that were manually matched to an EU variable $v$. MRR is the averaged reciprocal rank across all EU variables in a dataset.

$HR = V_{hit}^n / V_{all}$ (1)

$RR(v) = 1/R_{hit}$ (2)

$MRR = \sum_{v=1}^{all} RR(v)/V_{all}$ (3)

Ties in ranking were resolved by using the median rank when calculating HR and MRR. For example, if 3 JP variables have identical similarity scores when matched with an EU variable and occupy positions 4 through 6 in the ranked list, they will all be assigned a rank of 5.

## 2.7. Experimental settings

### 2.7.1 Model comparison

We first compared the performance of the four individual NLP methods on the whole dataset. We then compared the performance of the Random Forest classifier and the best NLP method on the test sets from 50 trials (see section 2.5.2 Training and test datasets). We used paired t-tests to compare model performance over the 50 trials, with the null hypothesis stating that there is no significant difference in performance metrics between these two methods. Five performance metrics were utilized, including the top-30, 20, 10, and 5 HR (HR-30, HR-20, HR-10, HR-5) and MRR.

### 2.7.2 Feature importance analysis

To understand which features contributed most to the performance of the Random Forest model, we estimated feature importance using data from the 50 trials and the permutation importance method [52]. For each feature in a trained model, permutation importance estimates its contribution by randomly shuffling the feature's values in a held-out dataset and measuring the subsequent decline in model performance on this dataset. In this study, we estimated feature importance by averaging the model's performance decline over the test sets from the 50 trials, measured by HR-5, HR-10 and MRR. We used these three metrics because we found that they were more sensitive to permutation on individual features in the Random Forest classifier, compared with HR-20 and HR-30.

We compared feature importance using two methods. In the first method, we ranked features using their mean importance scores over 50 trials. In the second method, we first ranked the features within each trial based on their importance scores, and then compared the average ranks across the 50 trials.
In addition, we conducted feature ablation analysis to assess whether removing a specific type (i.e., subset) of features from the Random Forest model would affect the model performance. We categorized the features into three types: LLM-derived features, fuzzy matching-derived features, and other features. We measured the average decline in model performance after removing each type of feature over 50 trials.

## 2.8 Error analysis

To understand the limitations of our variable-matching approach, we manually checked the manually matched EU-JP variable pairs that were ranked low (below top-30) by the best NLP method and the Random Forest model and summarized the error patterns.

## 3 Results

### 3.1 Descriptive statistics

The dataset used in this study included 352 GERAS-EU variables and 1322 GERAS-JP variables. As shown in Table 2, compared with the JP variables, the EU variables have longer labels (11.5 vs. 8.3 words), shorter data sheet descriptions (3.2 vs. 5.5 words), and longer derivation rules (13 vs. 10.8 words).

**Table 2. Characteristics of variable labels, data sheet descriptions, and derivation rules of EU and JP variables**

|  | EU | JP |
|---|---|---|
|  | n = 352 | n = 1322 |
| **Words in label** | 11.5 | 8.3 |
| **Words in data sheet description** | 3.2 | 5.5 |
| **Words in derivation rule** | 13.0 | 10.8 |

Appendix 2 provided descriptive statistics for features used in trial 1 of the machine learning experiment. Other trials showed similar patterns in feature value distributions. The NLP-derived features (i.e., similarity scores estimated by the individual NLP methods) showed similar distributions in both the training and test sets (Appendix 2, Table A2-1). A higher score or feature value indicates a greater similarity between the EU and JP variables.

Other features used in the Random Forest model were summarized in Table A2-2 in Appendix 2. The GERAS-EU study exhibits a shorter median keyword length (medium length: 9 words) in non-empty derivation rules than the GERAS-JP study (medium length: 15 words). The EU variables had longer labels (medium length: 10 words) than JP variables (medium length: 7 words). Approximately one-third of EU variables lacked the derivation rules, while only 11% of JP variables lacked the derivation rules.

### 3.2 Comparison of individual NLP methods

As shown in Table 3, the E5 model achieved the highest performance across all evaluation metrics, including a HR-30 of 0.898 and an MRR of 0.700. The MiniLM and MPNet models demonstrated slightly lower performance compared to the E5 model, with MiniLM outperforming MPNet across all evaluation metrics. The fuzzy matching method exhibited the lowest performance in most metrics but outperformed the MPNet model in HR-5 and MRR.

**Table 3. Performance of individual NLP methods in variable matching** [a, b]

| Models | HR-5 | HR-10 | HR-15 | HR-20 | HR-30 | MRR |
|---|---|---|---|---|---|---|
| **E5** | **0.798** | **0.861** | **0.886** | **0.889** | **0.898** | **0.700** |
| **MiniLM** | 0.662 | 0.730 | 0.778 | 0.795 | 0.841 | 0.551 |
| **MPNet** | 0.608 | 0.705 | 0.733 | 0.761 | 0.798 | 0.511 |
| **Fuzzy match** | 0.616 | 0.702 | 0.733 | 0.744 | 0.776 | 0.549 |

[a] The individual models were evaluated on all EU variables (n = 352) that were manually matched with JP variables

[b] HR-30: Top 30 hit ratio; HR-20: Top 20 hit ratio; HR-10: Top 10 hit ratio; HR-5: Top 5 hit ratio; MRR: Mean reciprocal rank; E5: E5_Large_V2 model; MiniLM: All_MiniLM_L12 model; MPNet: All_MPNet_base_V2 model; fuzzy matching: Token set ratio method. All methods used only variable labels to estimate variable similarities

### 3.3 Random Forest model versus E5

As shown in Table 4, the Random Forest model outperformed the E5 model on all the evaluation metrics and achieved an HR-30 of 0.977 (std: 0.015) and an MRR of 0.725 (std: 0.042). Paired t-tests on the 50 trials indicated significant performance gains (p < 0.001 for all metrics).

**Table 4. Random Forest and E5 model performance comparison**

| Metric [a] | E5 [b] mean (standard deviation) | Random Forest [b] mean (standard deviation) | Mean difference [b] (Random Forest – E5), 95% confidence interval | P value |
|---|---|---|---|---|
| **HR-30** | 0.911 (0.028) | 0.977 (0.015) | 0.066 [0.057, 0.075] | <0.001* |
| **HR-20** | 0.905 (0.027) | 0.961 (0.017) | 0.057 [0.048, 0.066] | <0.001* |
| **HR-10** | 0.871 (0.034) | 0.921 (0.028) | 0.049 [0.037, 0.062] | <0.001* |
| **HR-5** | 0.803 (0.040) | 0.863 (0.040) | 0.059 [0.044, 0.075] | <0.001* |
| **MRR** | 0.659 (0.044) | 0.725 (0.042) | 0.066 [0.049, 0.083] | <0.001* |

[a] HR-30: Top 30 hit ratio; HR-20: Top 20 hit ratio; HR-10: Top 10 hit ratio; HR-5: Top 5 hit ratio; MRR: Mean reciprocal rank; E5: E5_large_V2 model; RF: Random Forest model.

[b] The performances of E5 and Random Forest were compared over 50 trials, using the paired *t* test.

### 3.4 Feature analysis

As shown in Table 5, the feature derived by applying the E5 model on variable labels (i.e., E5_on_label) contributed most to the performance of the Random Forest model, as indicated by all the three-evaluation metrics. Other features differed in their contributions measured by HR and MRR. Features derived by applying MiniLM on variable labels (i.e., MiniLM_on_label) and by applying E5 on both variable labels and keywords extracted from the derivation rules (i.e., E5_on_label_key), and the label length of the JP variable (i.e., Label_len_JP) were among the top-ranked features for HR-5 and HR-10, followed by other LLM-derived features such as MPNet_on_sheet (for HR 5) and MiniLM_on_label_key (for HR-10). In contrast, features derived by applying LLMs on data sheet descriptions (e.g., MPNet_on_sheet, E5_on_sheet, MiniLM_on_sheet) and by applying fuzzy matching on variable labels (i.e., Fuzzy_on_label) were more important to MRR.

**Table 5. Feature importance for the Random Forest model [a]**

|  | HR-5 | | HR-10 | | MRR | |
|---|---|---|---|---|---|---|
| **Features** | **Average importance** | **Average rank** | **Average importance** | **Average rank** | **Average importance** | **Average rank** |
| **E5_on_label** | **0.132** | **1.0** | **0.106** | **1.0** | **0.127** | **1.0** |
| **MiniLM_on_label** | **0.025** | **5.7** | **0.026** | **5.0** | 0.012 | 7.9 |
| **E5_on_label_key** | **0.020** | **7.1** | **0.022** | **5.7** | 0.010 | 9.6 |
| **Label_len_JP** | **0.017** | **6.8** | **0.013** | **7.6** | 0.011 | 8.0 |
| **MPNet_on_sheet** | **0.015** | **7.3** | 0.011 | 8.3 | **0.015** | **6.9** |
| MPNet_on_label_key | 0.013 | 8.2 | 0.005 | 11.7 | 0.012 | 8.5 |
| MiniLM_on_label_key | 0.012 | 9.3 | **0.014** | **8.0** | 0.003 | 11.2 |
| Fuzzy_on_label | 0.012 | 9.3 | -0.003 | 14.0 | **0.014** | **7.6** |
| E5_on_sheet | 0.012 | 8.3 | 0.010 | 8.4 | **0.013** | **7.6** |
| MiniLM_on_sheet | 0.011 | 9.1 | 0.008 | 9.7 | **0.014** | **7.3** |
| Fuzzy_on_sheet | 0.010 | 9.3 | 0.007 | 10.3 | 0.009 | 9.4 |
| Derive_info_len_JP | 0.006 | 10.8 | 0.004 | 11.5 | 0.006 | 10.2 |
| Label_len_EU | 0.006 | 10.9 | 0.009 | 9.4 | 0.004 | 11.2 |
| Derive_info_len_EU | 0.002 | 13.0 | 0.001 | 12.5 | 0.001 | 12.4 |
| Derive_info_null_EU | 0.001 | 13.4 | 0.002 | 12.7 | 0.000 | 12.8 |
| Derive_info_null_JP | 0.000 | 14.0 | 0.000 | 13.6 | 0.000 | 12.7 |
| MPNet_on_label | 0.000 | 13.5 | 0.005 | 10.7 | -0.003 | 13.3 |
| Fuzzy_on_label_key | -0.001 | 14.0 | 0.005 | 10.9 | -0.005 | 13.4 |

[a] NLP-derived features are similarity scores calculated by three Large Language Models (E5, MPNet, MiniLM) and fuzzy matching using variable labels, data sheet descriptions, and key words extracted from the derivation rules (detailed in Section 2.4.3.2 Features and class label)

[b] HR-10: Top 10 hit ratio; HR-5: Top 5 hit ratio; MRR: Mean reciprocal rank

**Table 6. Feature ablation analysis [a]**

|  | Model 1: all features | Model 2: w/o LLM features | | Model 3: w/o fuzzy match features | | Model 4: w/o other features | |
|---|---|---|---|---|---|---|---|
|  |  | Δ (Model 1 – Model 2) [b] | *P* value | Δ (Model 1 – Model 3) [b] | *P* value | Δ (Model 1 – Model 4) [b] | *P* value |
| HR-30 | 0.98 [0.976, 0.984] | 0.07 [0.06, 0.08] | <0.001* | 0.002 [-0.004, 0.008] | 0.36 | 0.005 [-0.001, 0.011] | 0.014 |
| HR-20 | 0.96 [0.960, 0.971] | 0.10 [0.08, 0.11] | <0.001* | 0.003 [-0.005, 0.011] | 0.21 | 0.008 [-0.0003, 0.016] | 0.001* |
| HR-10 | 0.92 [0.916, 0.0.930] | 0.11 [0.10, 0.13] | <0.001* | 0.001 [-0.009, 0.012] | 0.63 | 0.017 [0.005, 0.028] | <0.001* |
| HR-5 | 0.86 [0.852, 0.874] | 0.15 [0.13, 0.17] | <0.001* | 0.009 [-0.007, 0.024] | 0.04* | 0.025 [0.009, 0.042] | <0.001* |
| MRR | 0.73 [0.712, 0.735] | 0.15 [0.13, 0.17] | <0.001* | 0.002 [-0.014, 0.018] | 0.56 | 0.014 [-0.003, 0.031] | <0.001* |

[a] HR-30: Top 30 hit ratio; HR-20: Top 20 hit ratio; HR-10: Top 10 hit ratio; HR-5: Top 5 hit ratio; MRR: Mean reciprocal rank

[b] The average performance drop of the partial model (i.e., model with features removed), compared to the full model

Feature ablation analysis (Table 6) showed that removing each type or subset of features decreased the model's performance in most cases (except for HR-10 when dropping the fuzzy matching features). The performance decline was greatest when removing LLM-derived features and was also notable when removing other types of features.

## 4. Discussion

Variable matching is a key early step in flexible data harmonization. We evaluated the performance of NLP methods, including LLMs and fuzzy matching, on this task. In addition, we developed and evaluated a Random Forest-based ensemble learning method that leveraged the strengths of individual NLP methods. We found that the E5 LLM outperformed other individual NLP methods on variable matching. The Random Forest model showed significantly better performance than E5 on all metrics and achieved an HR-30 of 0.98 and an MRR of 0.73. These results suggest that NLP techniques (including LLMs), combined with ensemble learning, have great potential in automating variable matching, thus speeding up the data harmonization process. Below, we discuss our main findings in greater detail.

Data harmonization has been discussed within the context of utilizing data from multiple sources, which aims to combine datasets for effective use by resolving data heterogeneity at three levels: syntax (i.e., data format), structure (i.e., conceptual schema), and semantics (i.e., how the variables were measured, derived, and encoded) [6]. The advantages of data harmonization include increasing the statistical power of a study and enabling big data analytics [6, 12, 53] verifying findings across studies [1], and evaluating and reducing the bias of analyses using individual data sources [2, 7]. Two strategies, merging and mapping, apply to the data harmonization process [6]. Merging involves the creation of a single global taxonomy or ontology for multiple datasets and then linking or mapping variables to the taxonomy [14, 54]. Mapping, on the other hand, involves the creation of a set of rules to match variables across studies [6]. For example, Kamala et al. harmonized data from two pregnancy cohort studies by using the mapping approach, where they created a set of rules by considering construct measured, question asked, response options, measurement scale, time and frequency of measurement, and coding features of variables, to classify variables into completely matching, partially matching (e.g., two variables measured the same construct but were measured or encoded in different ways), or no matching [12]. The mapping was conducted manually by reviewing documentation, consulting the research team of the original data source, and exploring variables in the dataset. Kamala's team harmonized 20 variables from both cohorts and pointed out that this is a repetitive and iterative process. In both mapping and merging approaches, matching variables between studies and the common data ontology or across studies can be time-consuming when performed manually but is a key step for data harmonization. Automated methods that facilitate variable matching, including the methods developed in this study, can potentially improve efficiency and, therefore, enable large-scale harmonization (e.g., harmonizing large datasets or many datasets).

In this study, we developed and tested new automated methods, leveraging LLMs and ensemble learning, to match variables across studies based on information provided in data dictionaries. Our approach relies on text comparison (e.g., comparing variable labels, data descriptions, and derivation rules) and, therefore, focuses on construct-level variable matching. Specifically, we considered both lexical similarity and semantic similarity of the text describing the variables [55]. The assessment of lexical similarity was motivated by the observation that matched variables sometimes have common technical terms in their definitions, such as variable labels or derivation rules. Fuzzy matching (approximate string matching [56]) focuses on measuring lexical similarity. The assessment of semantic similarity was motivated by the observation that matched variables could use different but semantically related words in their definitions. Text embedding is a widely used NLP technique for distributed text representation which can be used to measure semantic similarities between text [57]. Recent studies have shown the success of using sentence embeddings generated from BERT-based models on the semantic text similarity task [32, 37]. Large language models, like BERT, KeyBERT, and Roberta, also showed high performance in measuring short-text semantic similarities [58]. In this study, we treated variable matching as a short-text similarity assessment task and chose several pre-trained LLMs for measuring the semantic similarity between variable definitions. The evaluation of individual NLP methods showed that LLMs outperformed fuzzy matching, suggesting that measuring semantic similarities was beneficial for variable matching.

In addition, our evaluation results showed that the RF classifier consistently outperformed the best individual method (i.e., the E5 model). This is compatible with prior studies that showed that ensemble-based systems were more effective than single-expert systems [59]. In our case, the single-expert system, such as the fuzzy matching method or the LLMs, provides similarity scores for two variables. Due to variations in algorithm design and model training methods, these single-expert systems may offer diverse perspectives on the nuances of similarities and differences between variable definitions. Furthermore, ensemble learning provides a flexible framework for incorporating different information sources (e.g., variable labels, data sheet descriptions, and variable derivation rules in our case), which is also called data fusion [59]. In this study, we used the RF classifier to incorporate similarity scores measured by using different single-expert systems and information sources, as well as several additional features (e.g., lengths of variable labels), to classify each candidate GERAS-JP variable as matching or not matching a GERAS-EU variable. Results from feature importance analysis and feature ablation analysis showed that features derived by using LLMs contributed the most to RF's good performance. In addition, the feature importance analysis showed that the important features for a high hit ratio (which is sensitive to the quality of top-ranked variable pairs) and a high MRR (which measures the global or overall ranking quality) differed, although the E5-derived feature E5_on_label stood out as the most important for both metrics. The contribution of the Fuzzy matching feature seemed negligible in feature ablation analysis, except for the top-5 hit ratio. A possible reason is that LLM features already capture sufficient information about both lexical and semantic similarities for matching variables.

Through the error analysis, we observed two major error patterns. First, the description of a variable is sometimes ambiguous. For example, the variable "NPIRN" (labeled as "NPI item result numeric") from the GERAS-EU study failed to align with the variable "AVAL" (labeled as "analysis value") from the GERAS-JP study. Both variables represent the resulting number of NPI questions, but the label for the "AVAL" variable (i.e., analysis value) is too ambiguous or vague to be useful for variable matching. Second, the variable labels from the two studies sometimes emphasized different perspectives. For example, the variable "MMSES34" was labeled as "Item result numeric: Correct response to writing" in the GERAS-EU study, and its counterpart in the GERAS-JP study, i.e., MMSE_Q34, was labeled as "Please write a sentence". Both variables represent the evaluation result of item 34 in the MMSE questionnaire, but the variable label from the GERAS-EU study focused on the evaluation result, whereas the label from the GERAS-JP study focused on the evaluation item in the MMSE. In general, most error cases involved variable labels that lacked semantic and lexical similarities and, therefore, were difficult for the NLP methods we developed for variable matching.

We validated our approach on variable matching between the GERAS-JP and GERAS-EU studies. These two studies followed similar protocols to collect data, and their data dictionaries, which served as the inputs for our methods, were formatted similarly. For example, both data dictionaries include sections for variable name, variable label, data sheet description, and variable derivation rule. Additional work in data preprocessing is required when applying our methods to cases where data dictionaries of studies differ substantially in format and structure. However, the methodology underlying our approach is general. By utilizing pre-trained LLMs and the LLM-derived text similarity scores as machine learning features, our approach can be applied to other settings (e.g., ontology-based retrospective data harmonization, searching variables of interest in existing datasets or biobanks) where text descriptions about variables are available.

## 5. Conclusion

We found that NLP methods, which leveraged LLMs and ensemble learning, achieved promising results in the task of variable matching. Variable matching is an important early step in data harmonization, which often requires substantial human efforts. Our methods had great potential in reducing these manual efforts and can be applied where data dictionaries are provided for studies. In the future, we aim to apply and enhance our methods for scenarios where studies exhibit greater variation in data collection protocols.

**Fundings**

This study is supported by Alzheimer's Disease Data Initiative. JC was also supported by a 2023 pilot award from the Framingham Heart Study Brain Aging Program, which is funded by the National Institute on Aging (U19-AG068753), which also provides support to RA and AA. The American Heart Association (20SFRN35360180) provides funding support to RA, ZL, and ZP. The funders had no decisional role in study design, data collection and analysis, interpretation of data, or preparation of the abstract.

**Conflict of Interest**

RA is a scientific advisor to Signant Health and NovoNordisk. None of these potential competing interests overlap in any way with the content of the current study. The other authors have no conflict of interest.

**Ethics approval and consent to participate**

This study was conducted as a secondary data analysis of deidentified data made available through the AD Data Initiative's AD workbench and was exempt from IRB review.

**CRediT authorship contribution statement**

**Z.L:** Writing – review & editing, Writing – original draft, Software, Methodology, Data curation, Conceptualization, Formal analysis. **S.P.P:** Writing - Review & Editing, Software, Validation, Resources. **Z.T.P**: Writing - Review & Editing, Software, Validation, Resources. **S.S.J**: Writing - Review & Editing, Software, Validation, Resources. **V.B:** Writing - Review & Editing, Software, Validation, Resources. **T.F.A.A**: Writing - Review & Editing, Validation, Supervision, Funding acquisition. **R.A:** Writing – review & editing, Resources, Funding acquisition. **J.C:** Writing – review & editing, Writing – original draft, Supervision, Resources, Methodology, Conceptualization, Validation.

**Acknowledgments**

Data discovery services and computational resources contributing to this work were provided in kind by the AD Data Initiative [https://www.alzheimersdata.org/]. The authors would like to acknowledge the sponsor of the GERAS-EU study for making this data available upon request through AD Data Initiative's AD Workbench for analysis.

**Abbreviation**

AD: Alzheimer's Disease

BEIR: Benchmarking-IR

BERT: Bidirectional encoder representations from transformers

BMI: Body Mass Index

E5: EmbEddings from bidirEctional Encoder rEpresentations

EU: European Union

GERAS: A Prospective Observational Study of Costs and Resource Use in Community Dwellers with Alzheimer's Disease

HR: Hit Ratio

JP: Japan

LLMs: Large Language models

MPNet: Masked and Permuted Pre-training for Language Understanding

MiniLM: Deep Self-Attention Distillation for Task-Agnostic Compression of Pre-Trained Transformers

MRR: Mean Reciprocal Rank

MMSE: The Mini Mental State Examination

MTEB: Massive Text Embedding Benchmark

NLI: Natural Language Inference

NQ: Natural Questions

NLP: Natural Language Processing

RF: Random Forest

STS: Semantic Textual Similarity

SBERT: Sentence-BERT

XLNet: Generalized Autoregressive Pretraining for Language Understanding

# Figures

**Figure 1**. Overview of the automated variable matching approach.

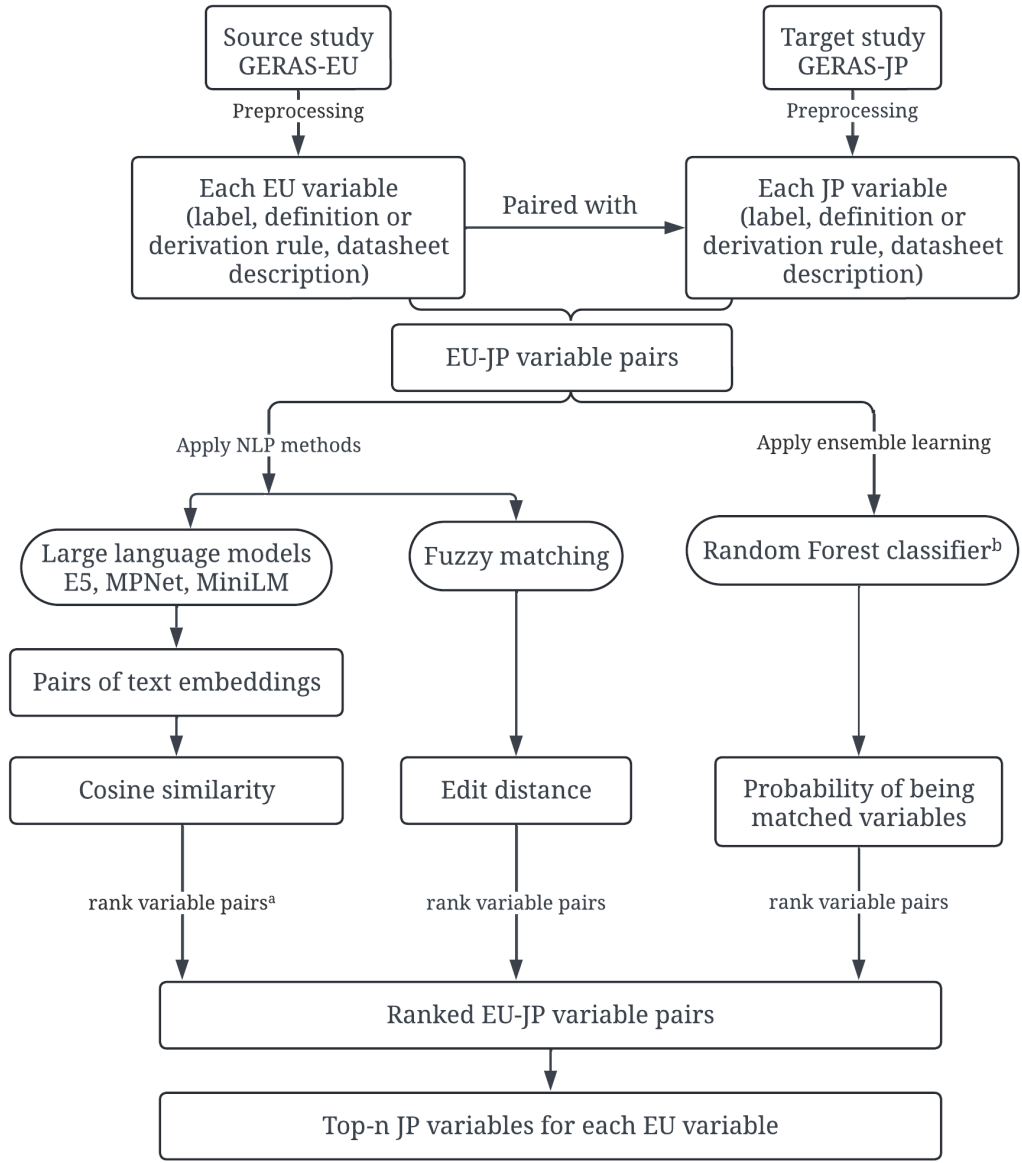

[a] Each large language model (LLM) was applied to rank EU-JP variable pairs separately

[b] The features used by the Random Forest classifier included cosine similarity scores generated by LLMs, edit distance scores generated by fuzzy matching, and other features derived from the data dictionary (detailed in section 2.3)

**Figure 2**. Training and evaluation of Random Forest (RF) classifier in a single trial.

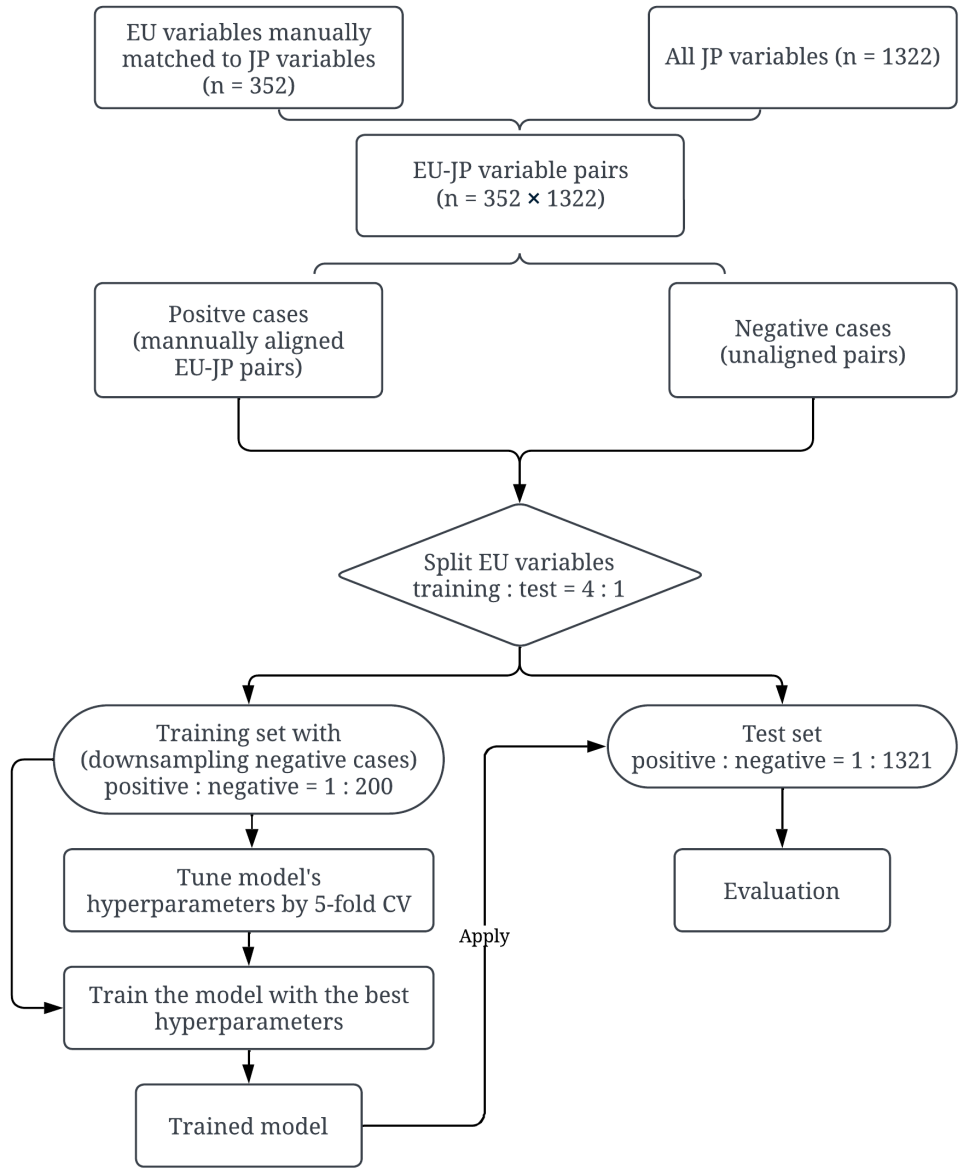

[a] The Random Forest classifier was trained and evaluated in 50 trials, with each trial having a different random split of the training and test datasets. Each test set had slightly varying ratios of positive (i.e., matched EU-JP variable pairs) to negative (i.e., unmatched EU-JP variable pairs) cases because some EU variables were manually aligned to multiple JP variables (see details in section 2.5.2).